# Hierarchical Deep Learning Ensemble to Automate the Classification of Breast Cancer Pathology Reports by ICD-O Topography

Waheeda Saib
IBM Research Africa
Johannesburg, South Africa
WSaib@za.ibm.com

David Moinina Sengeh
IBM Research Africa
Nairobi, Kenya
dsengeh@ke.ibm.com

Gciniwe Dlamini
IBM Research Africa
Johannesburg, South Africa
gciniwe.dlamini@za.ibm.com

Elvira Singh
National Cancer Registry
Johannesburg, South Africa
elvira.singh@nion.nhls.ac.za

## ABSTRACT

Like most global cancer registries, the National Cancer Registry in South Africa employs expert human coders to label pathology reports using appropriate International Classification of Disease for Oncology (ICD-O) codes spanning 42 different cancer types. The annotation is extensive for the large volume of cancer pathology reports the registry receives annually from public and private sector institutions. This manual process, coupled with other challenges results in a significant 4-year lag in reporting of annual cancer statistics in South Africa. We present a hierarchical deep learning ensemble method incorporating state of the art convolutional neural network models for the automatic labelling of 2201 de-identified, free text pathology reports, with appropriate ICD-O breast cancer topography codes across 8 classes. Our results show an improvement in primary site classification over the state of the art CNN model by greater than 14% for F1 micro and 55% for F1 macro scores. We demonstrate that the hierarchical deep learning ensemble improves on state-of-the-art models for ICD-O topography classification in comparison to a flat multiclass model for predicting ICD-O topography codes for pathology reports.

## 1 INTRODUCTION

While morbidity for non-communicable diseases is on the rise globally, cancer is particularly notable for its impact on both mortality and morbidity in Africa. In several sub-Saharan African countries, the most frequently diagnosed cancer among women is breast cancer, with South Africa attaining the highest rate of breast cancer incidence [7]. Cancer registries report data on diagnosed cancer cases and incidence rates. This is essential for healthcare resource and intervention planning. However, the processing and labelling of each new report submitted to a cancer registry is largely manual and time consuming, leading to notable time lags for reported and published data from registries. The South African National Cancer Registry (NCR), which is the most advanced cancer registry system in Africa has reported a 4-year lag for cancer registry data [2]. In the United States, there exists a 1.7-year delay in cancer reporting [8]. There is a clear need to develop methods that automate the processes used by registries, including report labelling using the International Classification of Diseases for Oncology (ICD-O), 3rd edition codes.

Several automated approaches have been proposed for labelling cancer reports. In [9], a literature review is provided of rule-based information retrieval (IR) and natural language processing (NLP) techniques. Rule-based systems were not generalizable across cancer domains and struggle with variability in report structure. While the author suggests machine learning (ML) approaches may be necessary for "complex and variable types of information", they conclude there is a "strong preference for rule-based systems over "black box" ML models in clinical practice." As the field of ML and deep learning (DL) applied to natural language processing (NLP) advances, there is increased explainability of black box models that outperform rule-based solutions [10]. A recent publication, [11] demonstrates that convolutional neural networks (CNNs), using word embeddings, consistently perform better than classical ML approaches using term frequency-inverse document frequency (TF-IDF) for IR and classification of pathology reports by primary tumor site. ICD-O codes are used for classification of cancers into topographical code (anatomical site of origin of the tumor), and morphological code (cell type of tumor or histology and behavior, i.e., malignant or benign). ICD-O classification is hierarchical in nature, with the top layer representing categories and the sub layers representing narrowing disease paths.

We thus use this knowledge as a basis to explore hierarchical CNNs for ICD-O classification of cancer reports. Hierarchical classification methods with conventional machine learning models such as SVM and Naıve Bayes, were used to create hierarchical tree structures of expert binary classifiers that solve a multiclass problem in [12]. Another approach diverges to explore a hierarchy of sub level multiclass classifiers, to attain better performance on multiclass problems [5]. Work in hierarchical text classification in deep learning, explores the different combinations of specialized deep neural networks, recurrent neural networks and CNNs as the first and second layer models for document classification [6]. With respect to hierarchical classification in the medical domain, [4] employs a type of hierarchical mixture of experts' approach that uses neural networks as the parent classifier and sub level linear classifiers such as Widrow-Hoff and Exponential Gradient to classify Medline medical abstracts. To the best of our knowledge, this work is the first to apply a hierarchical deep learning classification method based on specialized multiclass and binary CNN models to classify pathology reports. Our hypothesis is that for NLP tasks such as ICD-O topographical code classification that are hierarchical in nature, a combined approach to identify the class hierarchy and validate it using expert knowledge will achieve better performance than a flat multiclass model for classification of free text pathology reports. This could be potentially useful for deployment in a national cancer registry like NCR where the pathology reports cover a wide range of cancer types across different languages.

## 2 EXPERIMENTAL AND COMPUTATIONAL DETAILS

### 2.1 Data Description and Preparation

The data comprises 2201 anonymized breast cancer pathology reports from the NCR database. Each report has one of 9 ICD-O topography codes namely, C50.0, C50.1, C50.2, C50.3, C50.4, C50.5, C50.6, C50.8 and C50.9, with each code representing a breast cancer subtype or class. The report data is written in English, Afrikaans and a mixture of both. Of the 9 ICD-O breast cancer classes we use the 8 that have greater than 200 reports per class thereby excluding C50.6. We extract the report text from extensible markup language (XML) format. The Afrikaans only reports, numbers, stop words and punctuation symbols are removed from the report text which is then converted to lowercase and tokenized. We use the top 1400 TF-IDF features to filter the text of all pathology reports. This filtering reduces the length while retaining the significant words in each document. The dataset split employed was 80/10/10 for the train, validation and test set. Our reported results are on the unseen test set.

### 2.2 Experimental Design and Workflow

To simplify the complexity of the 8 ICD-O code classification problem, we decompose it into a hierarchical classification model, in the form of a two-layer tree structure, composed of CNN classifiers. To establish a baseline for model comparison and identify the ICD-O classes that should be separated into intermediate child classifiers, we assess the performance of a flat multiclass CNN model on all 8, 7 and 6 ICD-O classes, each time removing classes that are general in nature namely C50.8, and thereafter both C50.8 and C50.9. In addition to this, we analyze the confusion matrix to infer the classes that should be applied to separate classifiers as in [3] and the type of classification to use, that is, multiclass versus binary.

### 2.3 CNN for Text Classification

Convolutional neural networks have recently become the standard to benchmark text classification tasks. In NLP tasks, the input is a document matrix, where each row corresponds to a word density vector. CNNs are made up of the convolutional and pooling layers which act as a key feature generation mechanism. Convolutions are performed by applying a learned filter $w$ to a receptive field, described in terms of a sliding window of $h$ words. By applying the learned filter on every instance, on the $h$ word sliding window of the document results in feature maps that capture relevant properties of the words within each window. The feature maps are subsampled by taking the maximum values per dimension over different window results, in the subsequent pooling layer [14]. This encourages location invariance and decreases dimensionality of the output as it is passed to subsequent layers of the network. Through this mechanism, we learn key words and sequence phrases that contribute to the different categories in the ICD-O classification task.

### 2.4 Hierarchical Deep Learning

The contribution of this work, is our novel hierarchical CNN classification method applied to the prediction of ICD-O codes for cancer pathology reports. The multiclass classification problem is decomposed into a tree structure of specialized CNN classifiers that perform well on the identified class subsets. In this hierarchical method, illustrated by Figure 1, the parent classifier is a multiclass CNN designed to take as input all 8 ICD-O report classes and predict the group the report belongs to. The child classifiers are a multiclass and binary CNN model. A base CNN architecture for both multiclass and binary model is used, that randomly initializes an embedding layer with pathology words before applying the architecture as demonstrated by [13]. In this configuration, a convolutional layer computes 100 feature maps for each window of size 3, 4 and 5 words. Maxpooling is applied to the resulting feature map, and subsequently dropout of 0.5 is applied to this result. The result is then fed into a hidden layer before applying softmax classification. The model was trained for 147 epochs, 75 batch size using the adadelta optimizer.

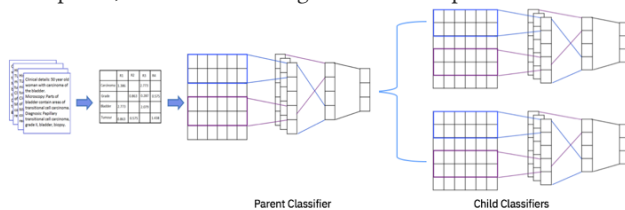

**Figure 1.** Hierarchical CNN Ensemble Layout

## 3 RESULTS AND DISCUSSION

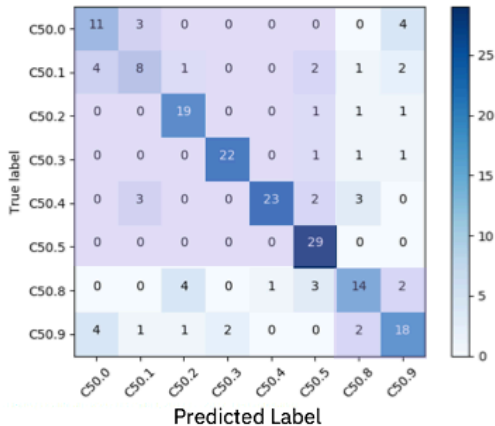

Figure 2: Confusion matrix of 8 class multiclass CNN illustrating one possible class grouping of 6 and 2 classes

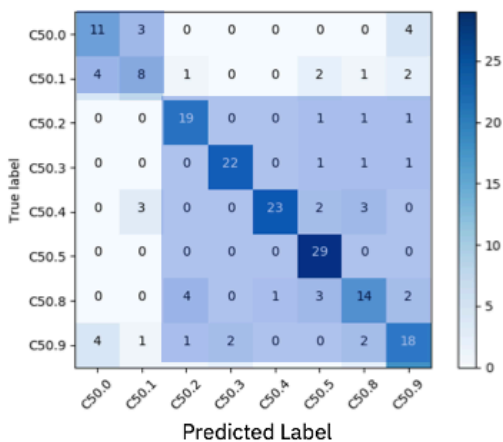

Figure 3: Confusion matrix of 8 class multiclass CNN illustrating an alternate class grouping of 6 and 2 classes

### 3.1 Class Relationship Analysis

From the confusion matrices of the flat multiclass CNN model in Figures 2 and 3, we see that C50.0, C50.1 and C50.8 are the most commonly misclassified classes. In Figure 2, the group of classes C50.0 to C50.5 are allocated to pathology reports that specify the location of the primary tumor site. Classes C50.8 and C50.9 are of a general nature namely, the former is allocated to pathology reports in which a lesion exists between two locations in the breast, and the latter is used when the location of the primary tumor is not specified in the pathology report. Hence, we remove C50.8 and C50.9 classes to observe performance on the more specific 6 class multiclass set and perform binary one-vs-one classification using C50.8 and C50.9. To compare an alternate grouping of classes, in Figure 3, we look at classes C50.0 and C50.1 as these are commonly misclassified and are both concerned with the primary site of the tumor in the central portion of the breast.

C50.0 and C50.1 are removed to view the performance on the 6 class multiclass with classes C50.2 to C50.9 and we perform binary one-vs-one on C50.0 and C50.1.

### 3.2 Comparative Analysis

We implemented 10-fold cross validation on the CNN model variations for different class subsets. During each fold, the model weights that performed well on the validation set were saved. Performance was measured using the F1-micro and F1-macro metric. The held-out test set during cross validation was used to evaluate the final model. The results were analyzed to understand the effect ICD-O classes had on overall performance. The results of the deep learning methods in Table 1, demonstrate the robustness of the CNN approach utilizing text filtering and random initialized word embeddings as it achieves good results on the different class experiments.

The previous state of the art for classification of breast cancer pathology reports by ICD-O topography on 7 classes, using CNNs with word embeddings, is reported by [11] to have an F1-micro of 0.644 and F1-macro of 0.213. Our flat CNN model results for 7 class ICD-O topography classification surpass these results by 14% for F1-micro and 55% for F1 macro.

We perform experiments to establish the best class groups of 6 and 2 classes to employ on the classifiers. Two possible class groups are illustrated in Figures 2 and 3. The predicted results from the cross-validation models were used to compute 95% confidence intervals. We report on the mean F1-micro and F1-macro of the models. Overall the best class grouping is shown by Figure 2, with the 6 class multiclass of C50.0 to C50.5 achieving a F1-macro of 0.826 and F1-micro of 0.846. Hence, this coupled with the expert logical definitions of the disease categories, we identify this as one of our specialized child classifiers in the hierarchical ensemble. With the general classes, C50.8 and C50.9, a binary one-vs-one approach is explored and optimized to attain results of F1-macro and F1-micro, both at 0.876. This is then used as the second specialized child classifier. For the parent classifier, the 8 classes are partitioned into two groups and applied to the CNN architecture with a final softmax layer. The best model for each classifier in the hierarchical ensemble was attained from cross validation. An automated pipeline was created to evaluate the hierarchical classification methods performance against the best flat multiclass model as illustrated in Figure 4.

```
ALGORITHM: Hierarchical CNN Classification
Group one ← Classes C50.8, C50.9
Group two ← Classes C50.0, C50.1, C50.2, C50.3, C50.4, C50.5
for each report in unseen report list, do
    group prediction ← First level CNN Multiclass ()   # predicts the group the report belongs to
    If group prediction in Group one
        ICD-0 class prediction ← One vs One CNN ()
    else
        ICD-0 class prediction ← Multiclass CNN ()
    Result list ← ICD-0 class prediction
return Result list
end
```

Figure 4: Automated Hierarchical Ensemble Evaluation

The results of the flat multiclass and the Hierarchical ensemble on the unseen data set are presented in Table 1. The performance of the 8 class multiclass CNN achieved an F1-macro and F1-micro score of 0.717 and 0.738 in comparison to the Hierarchical CNN ensemble which achieved F1-macro and F1-micro of 0.722 and 0.748. This demonstrates that for a multiclass problem, as the number of classes increase, so does the complexity. The performance increase in the Hierarchical ensemble can be attributed to it being composed of specialized deep learning models that have been optimized to perform well on the 6 and 2 class subsets.

**Table 1.** Comparison of Flat Multiclass and Hierarchical CNN

| ICD-O Classes | Model Variant | F1 Macro | F1 Micro | Accuracy |
|---|---|---|---|---|
| 8 | Multiclass CNN NPT | 0.703 (0.682, 0.723) | 0.724 (0.703, 0.744) | 0.724 (0.703, 0.744) |
| 7 | Multiclass CNN NPT | 0.769 (0.750, 0.788) | 0.792 (0.773, 0.811) | 0.792 (0.773, 0.811) |
| 6 C50.0-C50.5 | Multiclass CNN NPT | 0.826 (0.806, 0.846) | 0.846 (0.828, 0.865) | 0.846 (0.828, 0.865) |
| 2 C50.8, C50.9 | Binary OVO | 0.876 (0.849, 0.903) | 0.876 (0.85, 0.903) | 0.876 (0.85, 0.903) |
| 6 C50.2-C50.9 | Multiclass CNN NPT | 0.782 (0.763, 0.801) | 0.795 (0.775, 0.815) | 0.795 (0.775, 0.815) |
| 2 C50.0, C50.1 | Binary OVO | 0.777 (0.734, 0.819) | 0.778 (0.736, 0.822) | 0.778 (0.736, 0.822) |
| Final performance on test set | | | | |
| 8 | Multiclass CNN NPT | 0.717 | 0.738 | 0.738 |
| 8 | HECNN | 0.722 | 0.748 | 0.748 |

## 4  CONCLUSIONS

In this study, we propose a method to achieve better performance on a multiclass text classification problem. The method being to build a tree like structure of expert classifiers optimized to perform well on different class partitions. We demonstrate a way in which to determine the components of the hierarchical ensemble, by analyzing the class relationship between labels in the confusion matrix. This work diverges from the initial methods of hierarchical classification that decompose a multiclass problem into a tree of binary models. Initial results show that our presented hierarchical approach improves on a flat multi-class approach. Further exploration of this hierarchical model of classification will be employed across the entire ICD-O coding structure, which itself is hierarchical in nature. These results will further enhance the automatic classification of reports by NCR in the future.

### ACKNOWLEDGMENTS

This work was carried out through a Joint Research Study between IBM Research Africa and the National Cancer Registry in South Africa.